\documentclass[10pt,twocolumn,letterpaper]{article}

\usepackage{iccv}
\usepackage{times}
\usepackage{epsfig}
\usepackage{graphicx}
\usepackage{amsmath}
\usepackage{amssymb}
\usepackage[ruled,linesnumbered]{algorithm2e}
\usepackage{color,xcolor}
\usepackage{amsfonts,amssymb}
\usepackage{balance}
\usepackage{cancel}
\usepackage{ulem}
\usepackage{subcaption}
\usepackage{multicol}
\usepackage{multirow}
\usepackage{booktabs}
\usepackage{threeparttable}
\usepackage{bm}
\usepackage[breaklinks=true,bookmarks=false]{hyperref}

\iccvfinalcopy % *** Uncomment this line for the final submission

\def\iccvPaperID{****} % *** Enter the ICCV Paper ID here

\begin{document}

%%%%%%%%% TITLE
\title{A geometry-inspired decision-based attack}

\author{Yujia Liu\thanks{Work done during a summer internship at EPFL-LTS4.}\\
		University of Science and Technology of China\\
		Hefei, China\\
		{\tt\small yjcaihon@mail.ustc.edu.cn}
		\and
		Seyed-Mohsen Moosavi-Dezfooli\\
		\'Ecole Polytechnique F\'ed\'erale de Lausanne\\
		Lausanne, Switzerland\\
		{\tt\small seyed.moosavi@epfl.ch}
		\and
		Pascal Frossard\\
		\'Ecole Polytechnique F\'ed\'erale de Lausanne\\
		Lausanne, Switzerland\\
		{\tt\small pascal.frossard@epfl.ch}
	}

\maketitle
%\thispagestyle{empty}

%%%%%%%%% ABSTRACT
\begin{abstract}
Deep neural networks have recently achieved tremendous success in image classification. Recent studies have however shown that they are easily misled into incorrect classification decisions by adversarial examples. Adversaries can even craft attacks by querying the model in black-box settings, where no information about the model is released except its final decision. Such decision-based attacks usually require lots of queries, while real-world image recognition systems might actually restrict the number of queries. In this paper, we propose qFool, a novel decision-based attack algorithm that can generate adversarial examples using a small number of queries. The qFool method can drastically reduce the number of queries compared to previous decision-based attacks while reaching the same quality of adversarial examples. We also enhance our method by constraining adversarial perturbations in low-frequency subspace, which can make qFool even more computationally efficient. Altogether, we manage to fool commercial image recognition systems with a small number of queries, which demonstrates the actual effectiveness of our new algorithm in practice.
\end{abstract}

%%%%%%%%% BODY TEXT
\section{Introduction}

Deep neural networks have led to major breakthroughs in recent years and have been developing as powerful tools in various applications, including computer vision \cite{lecun2010convolutional, krizhevsky2012imagenet}, speech \cite{mikolov2011strategies, hinton2012deep}, health-care \cite{kriegeskorte2015deep, liang2014deep}, etc. Despite their huge success, it has been shown that these networks are vulnerable to deliberate attacks. An adversary can generate adversarial examples \cite{szegedy2013intriguing} that look very similar to the original images, yet that can mislead a deep neural network to give an incorrect output. 

Many existing attacks focus on white-box settings \cite{43405, carlini2017towards}, where adversaries have full access to all information of the model. They can attack models by computing gradients of the loss function. White-box attacks are interesting to derive a better understanding of deep neural networks, but they are unrealistic in a more practical setting, where adversaries have no knowledge about model parameters. In this case, called the black-box setting, some recent works focus on the score-based attacks \cite{narodytska2016simple, chen2017zoo}, where they make a large number of queries to the target model and exploit the output probabilities to generate adversarial examples. Most existing black-box attacks do however not consider that queries are usually costly in terms of both time and money, especially for commercial models. And, in fact, some commercial models only provide users with the final decision (top-1 label), and not even any output probabilities. This scenario corresponds to decision-based settings, where attacks \cite{brendel2017decision} are actually the most challenging for adversaries.

In this paper, we propose a novel method of decision-based attack called qFool that computes adversarial examples with only a few queries. We consider both non-targeted and targeted attacks, where we advantageously exploit the fact that the decision boundary is locally flat around adversarial examples. In non-targeted attacks, we estimate the gradient direction of the decision boundary only by relying on the top-1 label of each query result. We then search for an adversarial example directly from the original image in the estimated direction. In targeted attacks, we make iterative gradient estimations on multiple points along the boundary and seek for adversarial examples with the help of a target image. Both our non-targeted and targeted attacks can mislead networks with only few queries. Moreover, our experiments show that searching perturbations in a low-dimensional subspace is even more efficient in query-limited settings. We thus propose to extend qFool to subspaces in order to further reduce the number of queries. Finally, we use qFool to fool a famous commercial image recognition service with only a small number of queries to the model.

\vspace{0.2cm}
	\noindent Our main contributions are the following:
	 \begin{itemize}
	 \item We propose a novel method, qFool, for generating adversarial examples when only having access to the top-1 label decision of a trained deep network model. 
	 \item We demonstrate that qFool needs fewer queries than previous decision-based attacks in both non-targeted and targeted settings.
	 \item We enhance the proposed qFool with properly chosen subspace constraints, thus further reducing the number of queries while maintaining a high fooling rate.
	 \item We apply qFool to Google Cloud Vision, an example of commercially deployed black-box machine learning system. We demonstrate that such commercial classifiers can be fooled with a small number of queries.
	 \end{itemize}
%-------------------------------------------------------------------------
\section{Related Work}
Szegedy et al. \cite{szegedy2013intriguing} were the first to show the vulnerability of deep networks to adversarial examples.  Depending on how much adversaries know about networks, adversarial attacks are divided into white-box and black-box attacks. 

In white-box setting, adversaries have all the knowledge about the network itself.  It is a more favorable situation for adversaries to attack.
	
\vspace{0.1cm}
\noindent \textbf{Gradient-based attacks.}
\  Most white-box attacks need the gradient of loss function of the network. Goodfellow et al. \cite{43405} proposed a fast gradient sign method (FGSM), which explores the gradient direction. The Jacobian saliency map attack (JSMA) by Papernot et al. \cite{szegedy2016rethinking} uses the saliency map to compute the adversarial perturbation. DeepFool by Moosavi-Dezfooli et al. \cite{moosavi2016deepfool} generates an adversarial example by exploring the nearest decision boundary and crossing it to deceive the network. C\&W attack by Carlini and Wanger \cite{carlini2017towards} solves a more efficient optimization problem under three distance measurements with the Adam optimizer. Houdini by Cisse \cite{cisse2017houdini} is an approach for generating adversarial examples specifically tailored for task losses, and it can be applied to a range of applications.  Baluja and Fischer \cite{baluja2017adversarial} trained Adversarial Transformation Networks to generate adversarial examples with a very large diversity against a target network or set of networks. 
	
All these white-box attacks need to compute gradients of models, but sometimes attackers might have no access to the model or it may contain non-differentiable operations. So black-box attacks have gained more attention recently.
	
In the black-box attacks, adversaries do not have knowledge of the network. They only have access to the predictions by querying it. Black-box attacks can roughly be divided into three families: transfer-based, score-based and decision-based attacks.
	
\vspace{0.1cm}
\noindent \textbf{Transfer-based attacks.}
\ Szegedy et al. \cite{szegedy2013intriguing} found that adversarial examples can transfer between models, thus enabling black-box attacks on deployed models. Transfer-based attacks aim to train a surrogate model by exploiting predictions from an underlying target model. Papernot et al. \cite{papernot2017practical} showed that adversarial examples crafted based on surrogate models can mislead the target model. Liu et al.  \cite{liu2016delving} developed an ensemble transfer-based attack and showed its high success against Clarifai.com, a commercial image classification service.

\vspace{0.1cm}	 
\noindent \textbf{Score-based attacks.}
\ Adversaries only rely on the predicted scores of a model to generate adversarial examples in score-based attacks. Narodytska et al. \cite{narodytska2016simple} proposed the local-search attack that measures a model’s sensitivity to individual pixels. Chen et al. \cite{chen2017zoo} proposed Zeroth Order Optimization (ZOO), that approximates gradient information by the symmetric difference quotient to generate adversarial examples. Hayes et al. \cite{hayes2017machine} used the predicted scores to train an attacker model for black-box attacks.

\vspace{0.1cm}	
\noindent \textbf{Decision-based attacks.}
\ Decision-based attacks rely only on the class decision (top-1 label) of the model.  A simple method is the additive Gaussian noise attack \cite{rauber2017foolbox}, which probes the robustness of a model to i.i.d. normal noises. Adversaries can also use uniform noise, salt and pepper noise, contrast reduction or Gaussian blur. But perturbations computed by these simple methods are usually very perceptible. Brendel et al. \cite{brendel2017decision} proposed the Boundary attack, that can generate much smaller adversarial perturbations effectively. It starts from a large adversarial perturbation and iteratively reduces the norm of the perturbation by making the perturbed data point walk along the decision boundary while ensuring that it stays in the adversarial region. 
	
Most black-box attacks usually require a large number of queries to the network model, while this is an important constraint in practice.  Li et al. \cite{li2018query} introduced an active learning strategy to significantly reduce the number of queries for transfer-based attacks. For score-based attacks, Ilyas et al. \cite{ilyas2017query} applied natural evolution strategies and only used two to three orders of magnitude fewer queries than previous methods. Bhagoji et al. \cite{bhagoji2018practical} proposed random feature grouping and principal component analysis, which can achieve both high query efficiency and high success rate. Finally in decision-based attacks, since adversaries get less information from each query, the required number of queries is definitely large. Reducing the number of queries in decision-based is definitely challenging and in line with practical settings. In this paper, we propose a novel decision-based attack, which greatly improves the query efficiency even with commercial black-box systems.

\section{Non-targeted decision-based attacks}
	
We describe now our new decision-based attack algorithm. We consider a trained model with parameters $\theta$ that can be represented as $f_{\theta}: \bm{x} \rightarrow y$, where $\bm{x} \in \mathbb{R}^{d}$ is an input normalized image and $y$ is the final decision of the model, \textit{i.e.}, the top-1 classification label.
	
We first consider the non-targeted attack problem, where an adversary computes an adversarial perturbation $\bm{v}$ to change the estimated label of an image $\bm{x}_0$, \textit{i.e.},  $f_{\theta}(\bm{x}_0+\bm{v})\neq f_{\theta}(\bm{x}_0)$. Furthermore, there shall be no perceptible difference between the perturbed and original image, \textit{i.e.}, the Euclidean norm $||\bm{v}||_2\leq \tau$ for some small $\tau$.
The adversarial example is constructed by making queries to the unknown deep network. Theoretically, the more queries, the more information that one can get about the target model, and the smaller the adversarial perturbation. We aim to use as few queries as possible such that the norm of the perturbation is smaller than a certain threshold $\tau$.

\subsection{Query-efficient design}
\label{subsection: basic-qFool}
The basic idea of our decision-based attack algorithm is the following. Fawzi et al. \cite{fawzi2016robustness} have shown that the decision boundary has a quite small curvature in the vicinity of adversarial examples. This indicates that some geometry properties of the decision boundary can be approximated in a similar way at different neighbour points around adversarial examples. We therefore exploit this observation to compute the adversarial perturbation $\bm{v}$. The direction of the minimal adversarial perturbation $\bm{v}$ for the input sample $\bm{x}_0$ is theoretically the gradient direction of the decision boundary at $\bm{x}_{adv}$. As we do not have full access to the classifier's structure in the black-box scenario, it is impossible to compute the direction directly. But, since the boundary is quite flat, the gradient of the classifier at point $\bm{x}_{adv}$ is almost the same as the gradient at other neighboring points on the boundary. Hence, the direction of $\bm{v}$ can be approximated well by $\bm{\xi}$, the gradient estimated at neighbour point $\mathcal{P}$. Then we can seek for an adversarial $\bm{x}_{adv}$ from $\bm{x}_0$ along the approximated direction $\bm{\xi}$. The basic logic is illustrated in Fig.~\ref{fig:nontargeted_idea}. More details about each step are given below.
	
\begin{figure}
	\includegraphics[width=7.5cm]{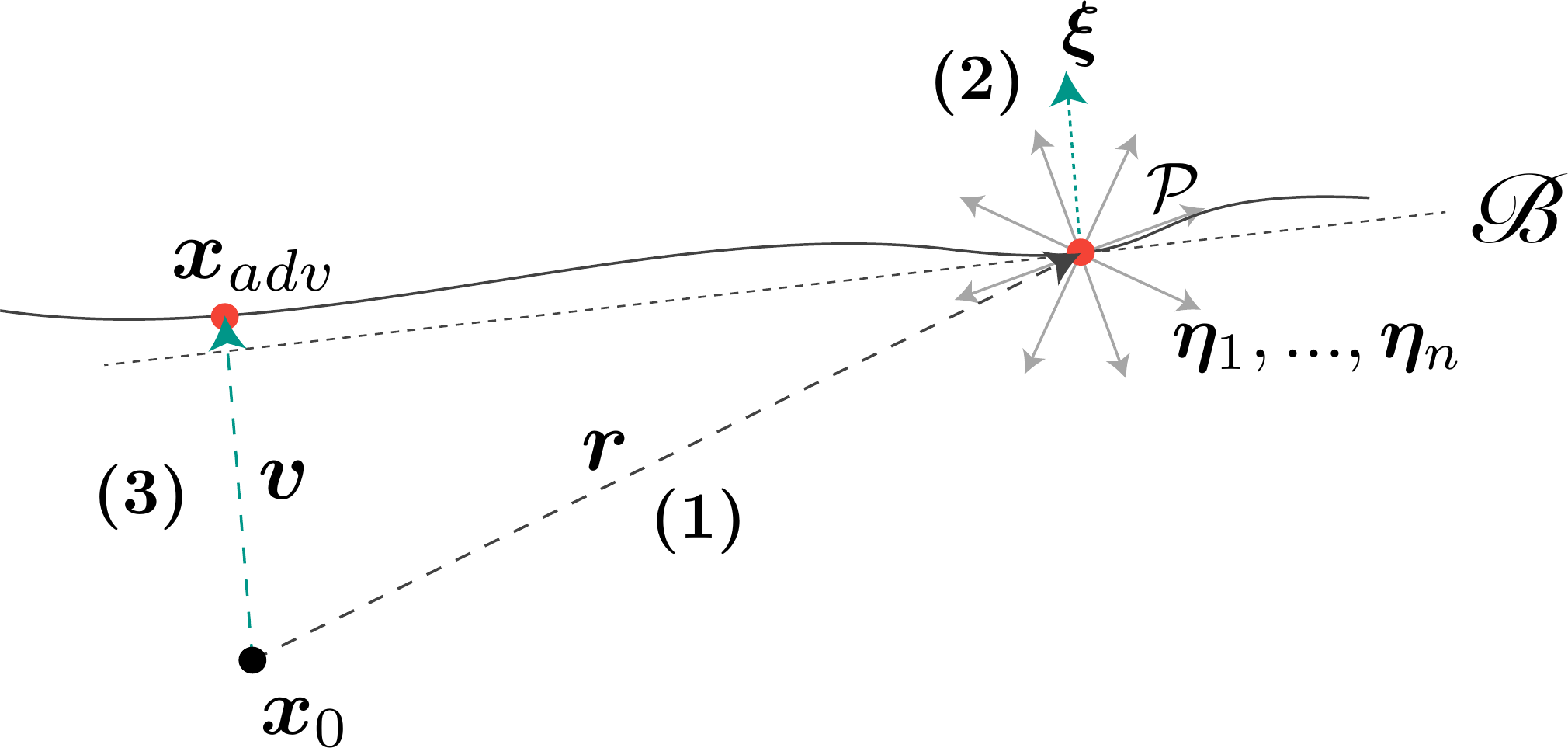}
	\vspace{-0.5em}
	\caption{A simple illustration of three steps to compute an adversarial example by qFool. (1) Compute starting point $\mathcal{P}$ with a random perturbation $\bm{r}$. (2) Estimate gradient direction $\bm{\xi}$ at $\mathcal{P}$ by $n$ perturbation vectors $\bm{\eta}_1,...,\bm{\eta}_n$. (3) Search an adversarial example $\bm{x}_{adv}$ with a perturbation $\bm{v}$ in the estimated gradient direction $\bm{\xi}$.} 
	\label{fig:nontargeted_idea} 
\end{figure}

\vspace{0.1cm}
\noindent \textbf{(1) Initial point.}
\ First, we find a small random noise $\bm{r}$ to perturb the original image $\bm{x}_0$ and identify a starting point $\mathcal{P}$ on the boundary. Formally,
\begin{equation}
	\mathcal{P}:=\bm{x}_0+\min_{\bm{r}}||\bm{r}||_2 \ \text{s.t. } f_\theta(\mathcal{P})\neq f_\theta(\bm{x}_0), \bm{r} \sim \mathcal{N}(0, \sigma)
	\end{equation}
To do that, we add some random Gaussian noises $\bm{r}_{i}\sim \mathcal{N}(0, \sigma_{i}) (i = 1,2,...,k)$ to the original image and make queries to the system, until the image $\mathcal{P}=\bm{x}_0+\bm{r}_j$ is misclassified. In order to make the starting point closer to the decision boundary, we then do a binary search, in the direction of $\bm{r}_j$ to find a small Gaussian noise $\bm{r}$ that makes the perturbed image located on the boundary. 
	
\vspace{0.1cm}
\noindent \textbf{(2) Gradient estimation.}
The gradient of boundary $\nabla f(\mathcal{P})$ is estimated using only the top-1 label of the classifier. We randomly generate $n$ small vectors $\bm{\eta}_{1}, ..., \bm{\eta}_{n}$ with the same norm to perturb $\mathcal{P}$ and query the classifier to get the predicted label $f(\mathcal{P}+\bm{\eta}_{i})$. We define:
\begin{equation}
z_{i}=
\begin{cases}
	-1 & f(\mathcal{P}+\bm{\eta}_{i})=f(\bm{x}_0)\\
	+1 & f(\mathcal{P}+\bm{\eta}_{i})\neq f(\bm{x}_0)
\end{cases}
,\ i=1,2,...,n
\label{zi}
\end{equation}

Theoretically, the vectors $(\bm{\eta}_{1}, ..., \bm{\eta}_{n})$ are likely to be symmetrically distributed on both sides of the decision boundary. For large enough $n$, $\frac{1}{n}\sum_{i=1}^n z_{i}\bm{\eta}_{i}$ converges to the normal of the decision boundary as the components of $\bm{\eta}_i$'s along the decision boundary will cancel out each other. Hence, the direction of the gradient $\nabla f(\mathcal{P})$ can be estimated as 
\begin{equation}
\bm{\xi}=\frac{\sum_{i=1}^n z_{i}\bm{\eta}_{i}}{||\sum_{i=1}^n z_{i}\bm{\eta}_{i}||_2}.
\label{estimate-direction}
\end{equation}
% where $||\cdot||_2$ is the Euclidean norm.

\vspace{0.1cm}
\noindent \textbf{(3) Directional search.}
Due to the flatness of the boundary, the direction of the gradient at point $\bm{x}_{adv}$ can be approximated by the one at point $\mathcal{P}$, \textit{i.e.}, $\nabla f(\mathcal{P})\approx \bm{\xi}$, which can be computed by Eq.~(\ref{estimate-direction}). We can thus find the adversarial example $\bm{x}_{adv}$ by perturbing the original image $\bm{x}_0$ in the direction of $\bm{\xi}$ until we hit the decision boundary. It only costs a few queries to the classifier by using a binary search algorithm. 
	
The qFool design described above is highly parallelizable, which can bring important speed-ups. In the step of gradient estimation, where the vast majority of queries are consumed, we can conduct parallel queries, as they are independent of each other. On the contrary, in all previous decision-based attacks, the algorithms are usually executed in an iterative way, and the results are highly dependent on the ones of the previous iterations. Hence, our algorithm can not only reduce the number of queries, but also save considerable computation time due to parallelization. 

\subsection{qFool algorithm}
\begin{algorithm}[ht]
	\caption{Non-targeted qFool}
	\label{alg: nontargeted_attack_n_algorithm}
	\KwIn{original image $\bm{x}_0$, initial estimation number $n_u$, a threshold $\epsilon$.}
	$i=0$ \;
	Search starting point $\mathcal{P}_0$ \;
	\While{$\sum_{j=0}^{i}{n^{(j)}}<n$} {
		${n}^{(i)} = 0$ \;
		$\bm{x}_{adv}^{(i)}= \bm{x}_{adv}' = \mathcal{P}_i$ \;
		\While{$\frac{\left\Vert \bm{x}_{adv}^{(i)}-\mathcal{P}_i\right\Vert}{n^{(i)}+n_u} \leq \epsilon \cdot\frac{\left\Vert \bm{x}_{adv}'-\mathcal{P}_i\right\Vert}{n^{(i)}}$ \textbf{ and } $\sum_{j=0}^{i}n^{(j)}<n$}{
		    Make $n_{u}$ new queries \;
		    $\bm{x}_{adv}' = \bm{x}_{adv}^{(i)}$ \;
		    $n^{(i)} = n^{(i)} + n_u$ \;
			Estimate gradient $\bm{\xi}_i$ at $\mathcal{P}_i$ with $n^{(i)}$ queries ~~~~~~in full/sub space\;
			Binary search for $\bm{x}_{adv}^{(i)} = \bm{x}_0 + \beta\cdot\bm{\xi}_i$ \;
		}
		$\mathcal{P}_{i+1}=\bm{x}_{adv}^{(i)}$, $i=i+1$  \;
	}
	\Return{$\bm{x}_{adv}^{(i-1)}$}
\end{algorithm}	

The distance between the starting point $\mathcal{P}$ and original point $\bm{x}_0$ directly impacts the quality of adversarial perturbations computed in Section~\ref{subsection: basic-qFool}. If we find a better starting point $\mathcal{P}$ that is closer to the boundary, the final perturbation is generally smaller and more likely to be imperceptible.
	
Therefore, we design the qFool attack as an iterative algorithm where the starting point $\mathcal{P}$ is iteratively updated. The total number of queries $n$ is divided into $s$ iterations, \textit{i.e.}, $n= n^{(0)}+n^{(1)}+...+n^{(s-1)}$. $s$ and $n^{(i)}$ $(i=0,...,s-1)$ can be found adaptively. At $i$-th iteration, we query the model $n^{(i)}$ times. These queries permit to estimate the gradient direction $\bm{\xi}_i$ at starting point $\mathcal{P}_i$, and to search for the adversarial example $\bm{x}_{adv}^{(i)}$ in the direction of $\bm{\xi}_i$. Then we update the starting point $\mathcal{P}_{i+1}$ to $\bm{x}_{adv}^{(i)}$ in the next iteration. 
	
The allocation of the number of queries $n= n^{(0)}+n^{(1)}+...+n^{(s-1)}$ in the gradient estimation plays an important role in the efficiency of the algorithm. Here we find $n^{(i)}$ $(i=0,1,...s-1)$ in a heuristic and greedy way. In one iteration, we first initialize the number of queries with a small $n_u$ and compute an adversarial perturbation following the three steps described in Section~\ref{subsection: basic-qFool}. Then we gradually increase the number of queries in this iteration to get a smaller perturbation. This process will stop when the reduction rate of the perturbation is less than a preset threshold or the number of used queries reaches $n$. With this method, we can find an appropriate number of queries for each iteration $n^{(i)} (i=0,1,...,s-1)$. 

The norm of noise vectors in the step of estimating gradient direction is another important parameter. It should be small enough so that the boundary between the adversarial and non-adversarial regions can be considered as approximately linear. If so, we expect to get an equal split of $+1$ and $-1$ for $z_{i}$ in Eq.~(\ref{zi}). We dynamically adjust the norm $\omega^{(k)}$ in the $k$-th iteration according to the norm and the percentage of $z_i=+1$ in previous iterations. We formalize this adjustment in Eq.~(\ref{eq: omega}).
\begin{equation}
\label{eq: omega}
	\begin{aligned}
	&\omega^{(k+1)}=\omega^{(k)}\cdot (1+\phi^{(k)}\cdot \rho^{(k)}) \\
	&\phi^{(k)}=-\text{sign}(\rho^{(k)})\cdot\phi^{(k-1)}
	\end{aligned}
	\vspace{-0.2em}
\end{equation}
where $\omega^{(0)}=\omega_{0}$, $\phi^{(0)}=-1$. $\phi^{(k)}$ controls the increase or decrease of $\omega^{(k)}$, and $\rho^{(k)}$ is the difference between 0.5 and the proportion of $z_i$ that are equal to $+1$. In this way, we can always ensure that the noise vectors are evenly distributed on both sides of the boundary.

\subsection{qFool in subspace}
\label{subsection: subspace-qFool}
In order to further reduce the number of queries, we now propose to concentrate on perturbations that belong to a properly chosen subspace, in order to simplify the search for adversarial perturbations. Indeed, it has been shown that adversarial perturbations that cause data misclassification can be found within specific subspaces \cite{fawzi2016robustness}. Hence, in the step of gradient estimation, instead of using $n$ noise vectors $\bm{\eta}_{1}, ..., \bm{\eta}_{n}$ of dimension $d$, we randomly choose $n$ noise vectors in a pre-defined subspace of dimension $m\ll d$. The other steps of the algorithm remain the same. The non-targeted qFool algorithm is described in Alg.~\ref{alg: nontargeted_attack_n_algorithm}.

Generating adversarial perturbations in subspaces rather than full data space clearly reduces the complexity of querying the model. The more symmetrically the sampled noise vectors are distributed, the better the components in the other directions than the gradient direction will cancel out, so the more accurate the estimated gradient direction. Given a limited number of sampled noise vectors, these vectors are distributed more sparsely in the full space than in a lower dimensional subspace. Thus, the probability that these vectors are symmetrically distributed on both sides of gradient direction in a subspace is larger than the one in the full space. This leads to a smaller adversarial perturbation in a subspace. Alternatively, given a threshold of the norm of perturbation, the subspace method requires fewer queries.
	
\section{Targeted attacks} 
We now extend our method to the targeted attack problem. In this case, an adversary tries to compute an adversarial perturbation $\bm{v}$ to change the label towards a specific target class $t$, \textit{i.e.}, $f_{\theta}(\bm{x}_0+\bm{v})= t$. The norm of $\bm{v}$ should be small enough. The only information available from the classification system is the top-1 label and the number of queries is assumed to be limited.

Similarly to the non-targeted attack, we need a starting point in the target adversarial class, close to the decision boundary. But it is almost impossible to change the label of the original image to a specific target label by adding random noise. We therefore utilize an arbitrary image $\bm{x}_t$ belonging to the target class $t$ to find a proper starting point $\mathcal{P}$. 

\begin{figure}[t]
	\includegraphics[width=8cm]{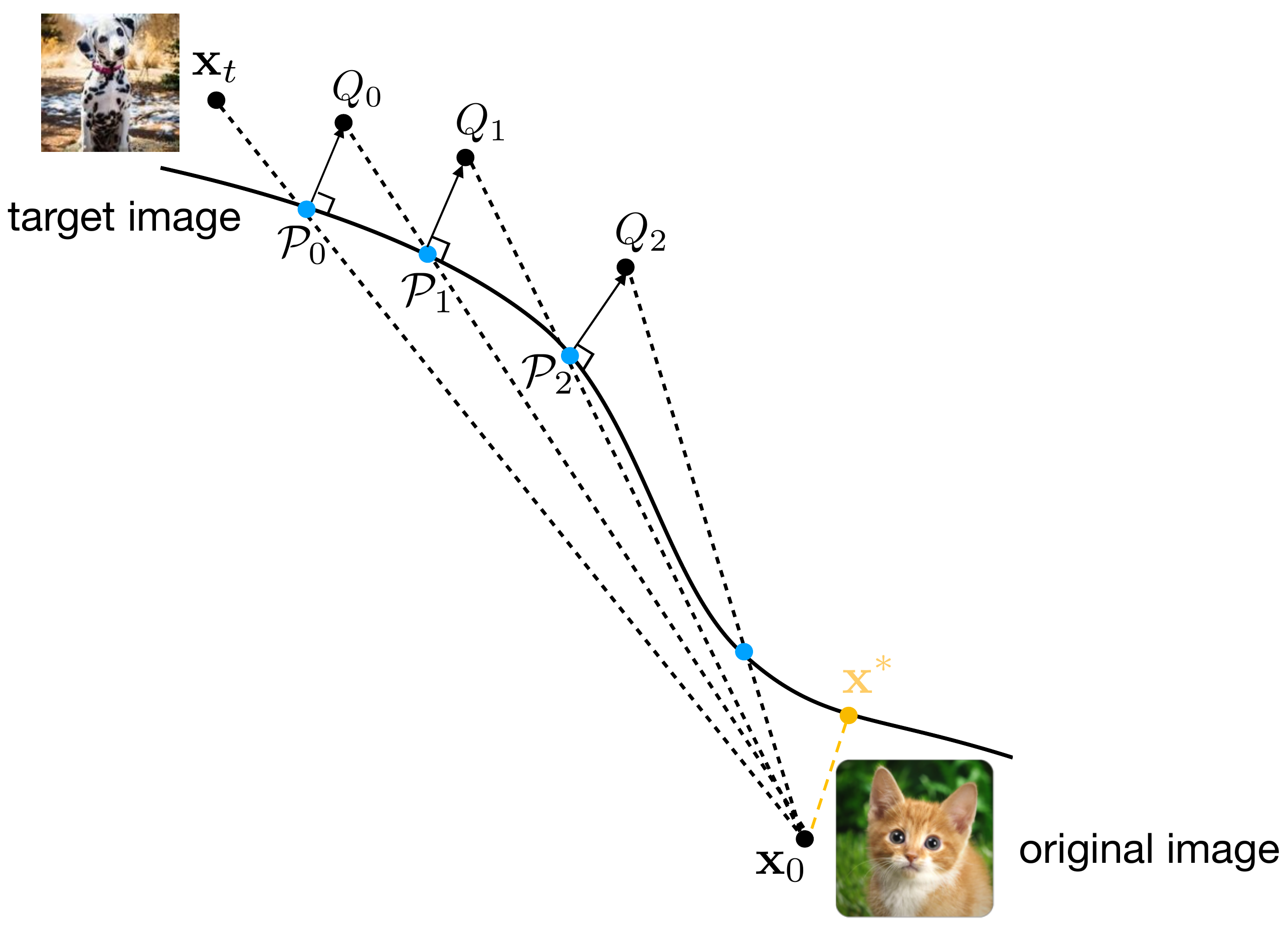}
	\vspace{-0.5em}
	\caption{An illustration of qFool in targeted attacks.} \label{fig:targeted_idea}
\end{figure}

As the distance between $\bm{x}_0$ and $\bm{x}_t$ can be quite large, the decision boundary between original and target adversarial region cannot be treated as flat. We cannot find a good estimation around $\mathcal{P}$ as we did in the non-targeted attack case. In the targeted attack, we therefore use a new way to compute and update the starting point $\mathcal{P}_i$ in each iteration, as shown in Fig.~\ref{fig:targeted_idea}. We first perform a linear interpolation in the direction of ($\bm{x}_{t}-\bm{x}_0$) to find a starting point $\mathcal{P}_0$,
\begin{equation}
\mathcal{P}_0:=\min_{\alpha}(\bm{x}_0+\alpha\cdot \frac{\bm{x}_t-\bm{x}_0}{||\bm{x}_t-\bm{x}_0||_2}) \quad \text{s.t. }f_\theta(\mathcal{P}_0)= t
\label{eq: targeted-P}
\end{equation}

\begin{algorithm}[htb!]
	\caption{Targeted qFool}
	\label{alg: targeted_attack_algorithm}
	\KwIn{original image $\bm{x}_0$, target image $\bm{x}_t$, initial estimation number $n_u$, a threshold $\epsilon$.}
	$i=0$ \;
	Compute direction $\bm{\nu}_0=\frac{\bm{x}_t-\bm{x}_0}{||\bm{x}_t-\bm{x}_0||_2}$ \;
	Search starting point $\mathcal{P}_0$ in direction $\bm{\nu}_i$ by Eq.~(\ref{eq: targeted-P}) \;
	\While{$\sum_{j=0}^{i}n^{(j)}<n$} {
		${n}^{(i)} = 0$ \;
		$\bm{x}_{adv}^{(i)}= \bm{x}_{adv}'= \mathcal{P}_i$ \;
		\While{$\frac{\left\Vert \bm{x}_{adv}^{(i)}-\mathcal{P}_i\right\Vert}{n^{(i)}+n_u} \leq \epsilon \cdot\frac{\left\Vert \bm{x}_{adv}'-\mathcal{P}_i\right\Vert}{n^{(i)}}$ \textbf{ and } $\sum_{j=0}^{i}n^{(j)}<n$}{
		    Make $n_{u}$ new queries \;
		    $\bm{x}_{adv}' = \bm{x}_{adv}^{(i)}$ \;
		    $n^{(i)} = n^{(i)} + n_u$ \;
			Estimate gradient $\bm{\xi}_i$ at $\mathcal{P}_i$ with $n^{(i)}$ queries  ~~~~~~in full/sub space\;
			$Q_i=\mathcal{P}_i+\delta\cdot\bm{\xi}_i$ \;
			$\bm{\nu}_i=\frac{Q_i-\bm{x}_0}{||Q_i-\bm{x}_0||_2}$ \;
			Search $\bm{x}_{adv}^{(i)}$ in the direction $\bm{\nu}_i$ by Eq.~(\ref{eq: targeted-P}) \;
		}
		$\mathcal{P}_{i+1}=\bm{x}_{adv}^{(i)}$, $i=i+1$  \;
	}
	\Return{$\bm{x}_{adv}^{(i-1)}$}
\end{algorithm}

\begin{figure*}[htb!]
\centering
    \includegraphics[width=0.9\textwidth]{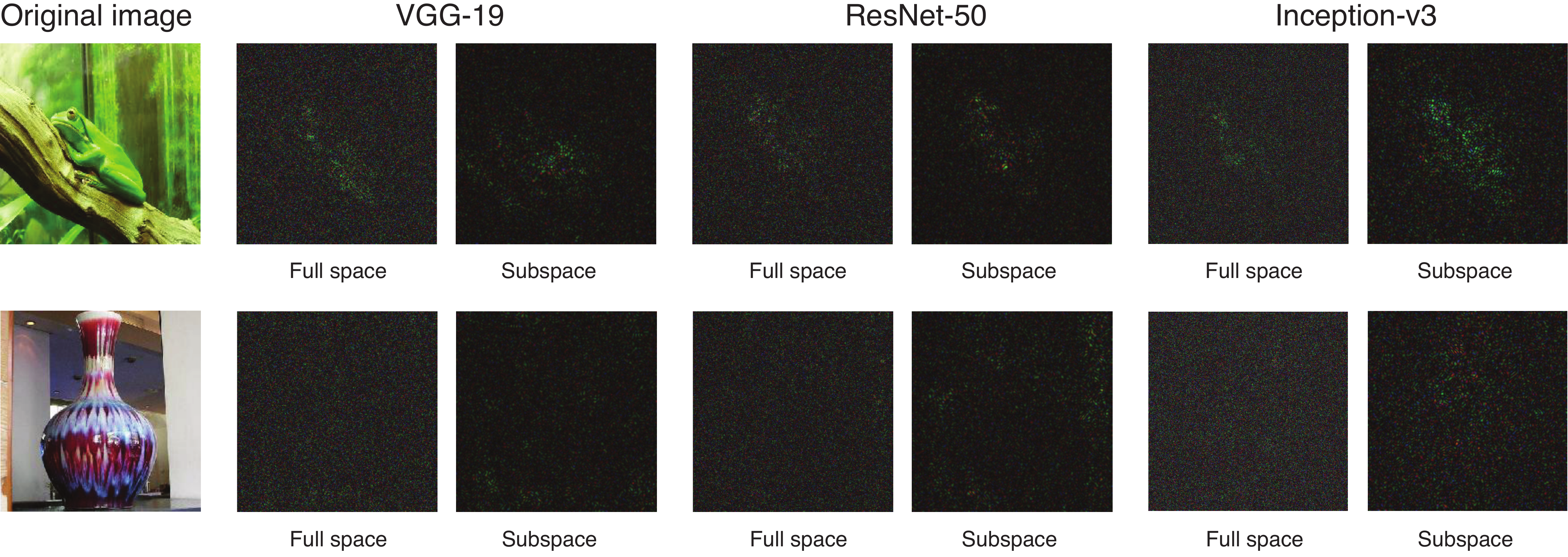}
    \vspace{-0.5em}
	\caption{Original images and adversarial perturbations by qFool in the full space and subspace on different models for a ``frog'' and a ``vase'' (20,000 queries). The resulting perturbed images are classified as ``snake" (first row) and ``goblet'' (second row). In addition, qFool in the subspace can get smaller perturbation than full space for the same number of queries.}
	\label{fig: nontargeted-perturbation-show}
\end{figure*}

\begin{figure*}[ht]
	\begin{minipage}[t]{0.32\textwidth}
		%\centering  
		\includegraphics[width=6cm]{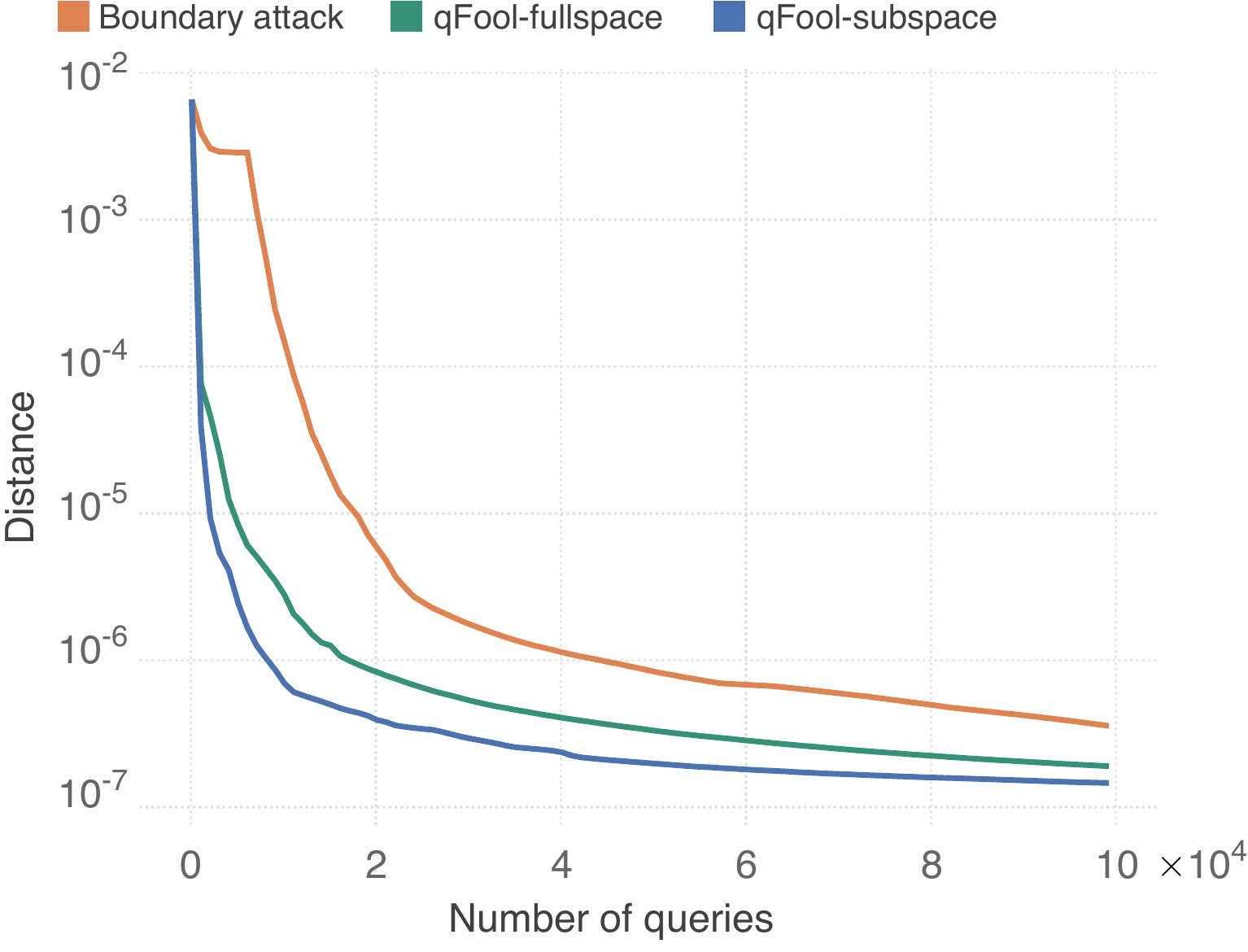}  
		\caption*{(a) VGG-19}
	\end{minipage} 
    \begin{minipage}[t]{0.32\textwidth}
    	%\centering  
    	\includegraphics[width=6cm]{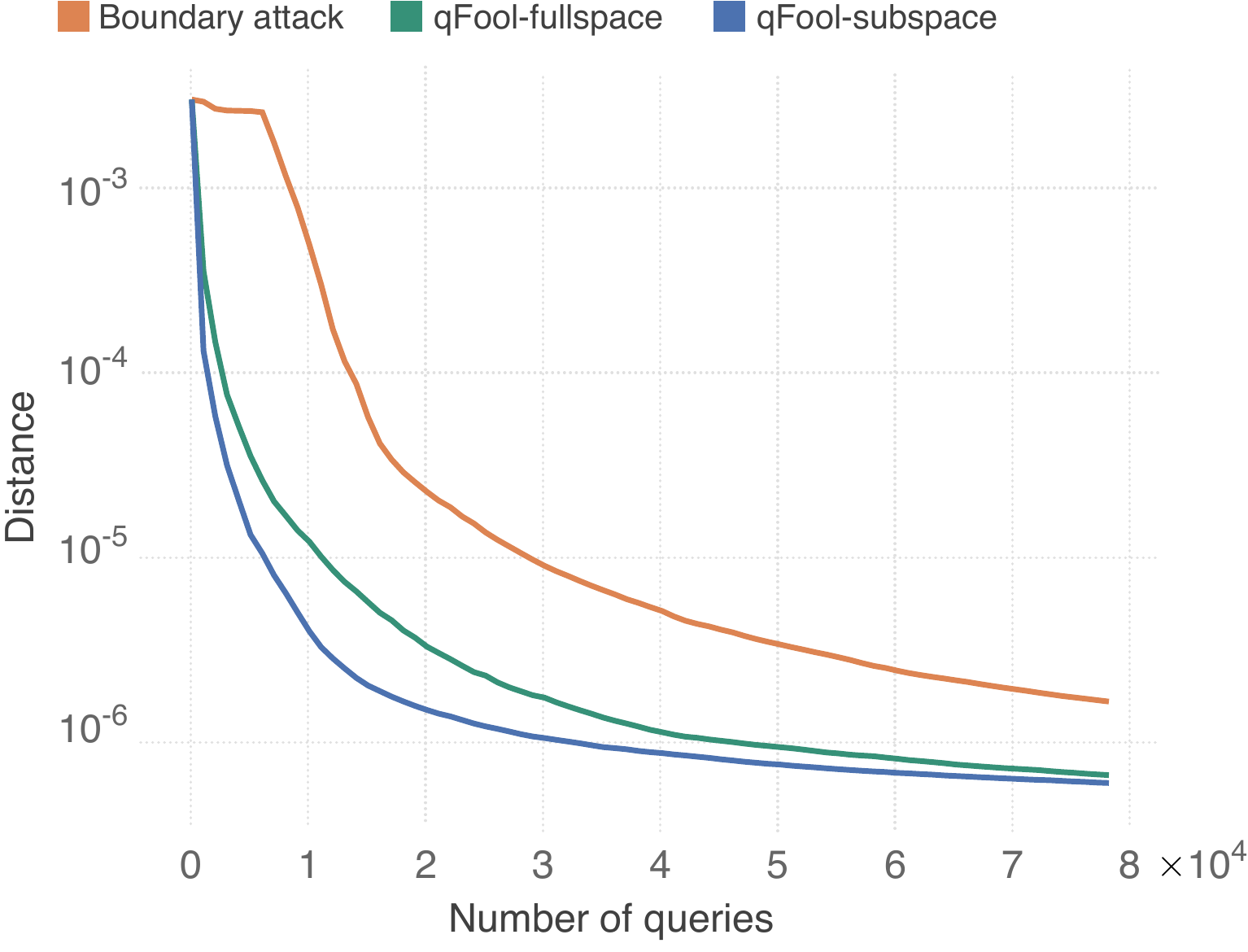} 
    	\caption*{(b) ResNet-50}
    \end{minipage} 
    \begin{minipage}[t]{0.32\textwidth}
	%\centering  
	\includegraphics[width=6cm]{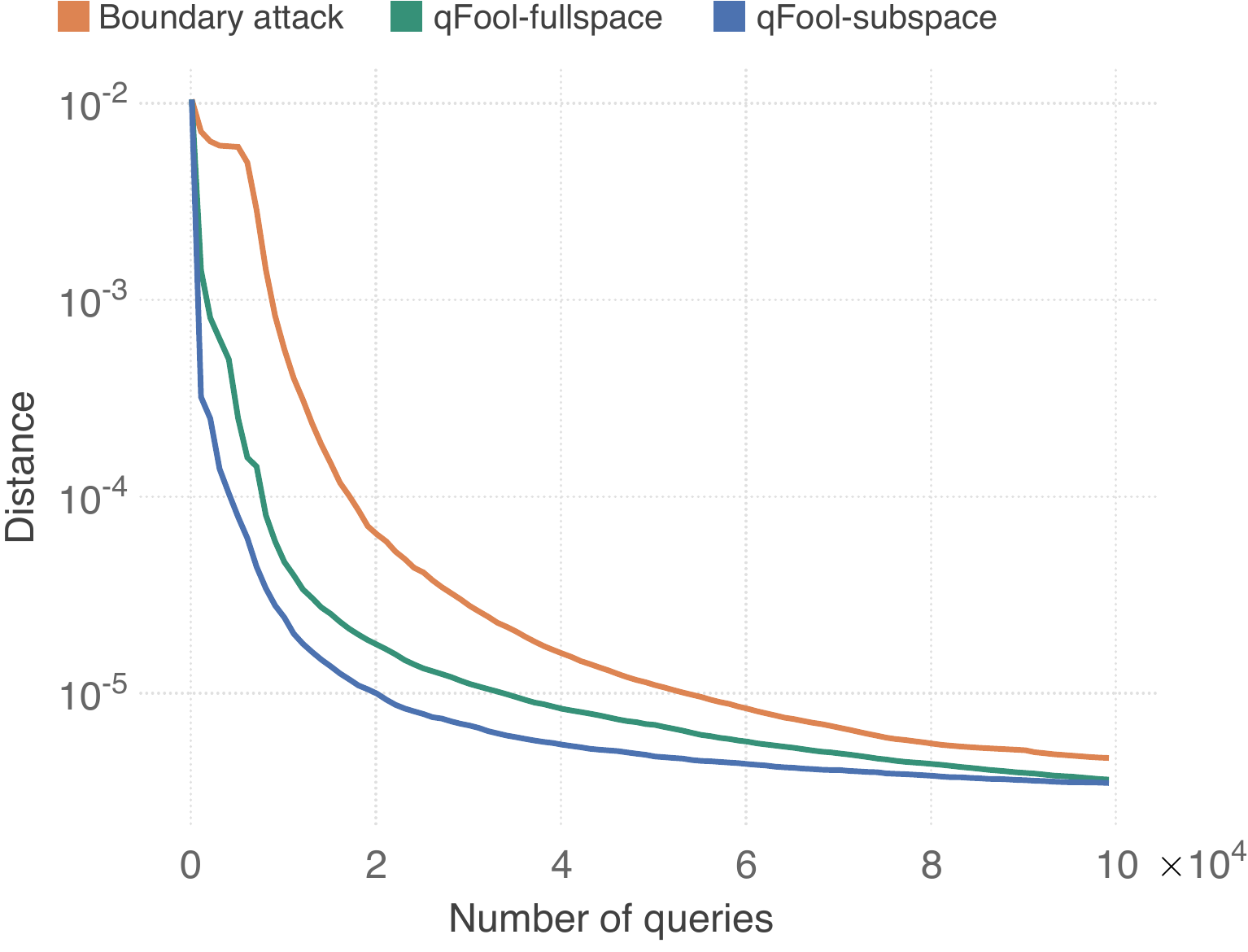}  
	\caption*{(c) Inception-v3}
\end{minipage} 
\vspace{-0.5em}
	\caption{Distance (the median of MSE in Eq.~(\ref{eq: MSE})) between adversarial example and original image over numbers of queries in non-targeted qFool and Boundary attack \cite{brendel2017decision} against different models.}
	\label{fig: nontargeted-boundary-full-sub}
\end{figure*}
The gradient direction $\bm{\xi}_0$ at $\mathcal{P}_0$ can be estimated with the same method as for non-targeted attacks. We then get a point $Q_0$, which is slightly away from the boundary, by adding a small perturbation at $\mathcal{P}_0$ in this direction, \textit{i.e.}, $Q_0=\mathcal{P}_0+\delta\cdot \bm{\xi}_0$. In the direction of ($Q_0-\bm{x}_0$), $\mathcal{P}_1$ can be searched with the method in Eq.~(\ref{eq: targeted-P}). Motivated by $\mathcal{Q}_{i}$ $(i=0,1,...,m)$, another point $\mathcal{P}_{i+1}$ near the boundary can be found quickly. Points $\mathcal{P}_{i}$ can walk along the boundary until the computed adversarial perturbation converges, \textit{i.e.}, $(\mathcal{P}_{m}-\bm{x}_0) \parallel \nabla f(\mathcal{P}_{m})$, and finally $\mathcal{P}_{m}$ is the adversarial example we get. Alg.~\ref{alg: targeted_attack_algorithm} describes the algorithm in details. We note that the targeted qFool attack algorithm can also be executed in either the full data space or limited to a pre-defined subspace.

\section{Experimental Results}
We evaluate qFool using 250 images randomly chosen from the validation set of ImageNet on the networks  VGG-19 \cite{simonyan2014very},  ResNet-50 \cite{he2016deep} and Inception-v3 \cite{szegedy2016rethinking}. We measure the median of the average norm of all adversarial perturbations after a specific number of queries, defined by

\begin{equation}
\mathcal{M}_{\mathcal{X}}(n)=\underset{\bm{x}_i \in \mathcal{X}}{\text{median}}\left(\frac{1}{m}\left\|\bm{v}(\bm{x}_i, n)\right\|_{2}^{2}\right),
\label{eq: MSE}
\end{equation}
where $\bm{v}(\bm{x}_i, n)\in \mathbb{R}^{m}$ is an adversarial perturbation generated for sample $\bm{x}_i$ using $n$ queries to the model. The median is taken over the images in dataset $\mathcal{X}$.

\subsection{Non-targeted attacks}

For the non-targeted attack, Fig.~\ref{fig: nontargeted-perturbation-show} shows samples of adversarial perturbations on different models. We compare qFool with Boundary attack \cite{brendel2017decision} on ImageNet. The results can be seen in Fig.~\ref{fig: nontargeted-boundary-full-sub}.  When the distances between adversarial examples generated by the two methods and original images are similar, qFool always spends fewer queries. qFool can converge much faster: if the number of queries is limited to a small value (\textit{e.g.}, 10000), our method can achieve much better performance.  

In addition, compared with qFool in full space, the subspace version can reduce the number of queries even more. Specifically, in this work, we use a 2-dimensional Discrete Cosine Transform (DCT) basis to define the low-dimensional subspace. Let $\mathcal{S}=\{\bm{\psi}_{i,j}\}_{i,j=0,...,\sqrt{m}-1}$ be the basis in the subspace of dimension $m$. When estimating gradients in the subspace, we use $n$ noise vectors $\bm{\eta}^{*}_{i}=\mathcal{S}\bm{\gamma}_{i}$, 
where $\bm{\gamma}_{i}\sim{\mathcal{N}(\boldsymbol{0},\boldsymbol{I}_m)}$, instead of vectors $\bm{\eta}_{i}\sim{\mathcal{N}(\boldsymbol{0},\boldsymbol{I}_d)}$.

\begin{figure}[ht]
	\includegraphics[width=0.4\textwidth]{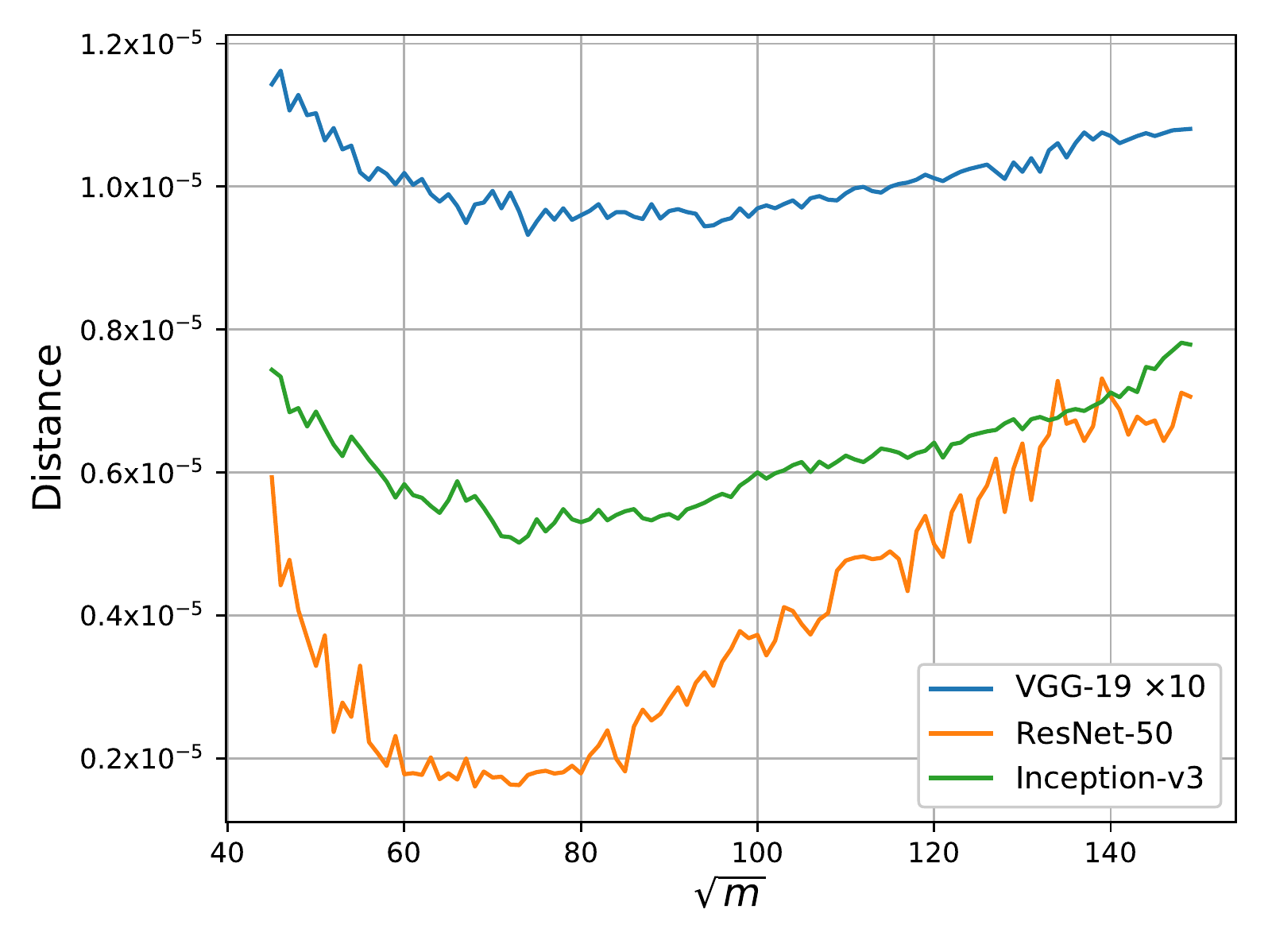}
	\vspace{-0.5em}
	\caption{$\ell_2$-distance between adversarial examples and original images in ImageNet training set over different subspace  dimensions in non-targeted qFool attacks against three classifier models. (Results for VGG-19 are multiplied by 10 to make the dynamic range roughly the same.)} 
	\label{fig: best-subspace}
\end{figure}

In our experiments, the dimension $d$ of full space is $224\times 224$ or $299\times 299$, we use 250 images in the ImageNet training set to find the best subspace. From Fig. \ref{fig: best-subspace}, the best dimensions $m$ of subspace range from $70\times 70$ to $90\times 90$. We choose $\sqrt{m}=75$, \textit{i.e.}, we use a low frequency subspace $\mathcal{S}$ of dimension $m=75\times 75$. The method of using perturbation vectors in the subspace to estimate the gradient of boundary can achieve higher efficiency, as shown in Fig.~\ref{fig: nontargeted-boundary-full-sub}. 

We also compare qFool with several white-box methods on 50 images when applying no more than 10,000 queries, shown in Table. \ref{tab:comparison-nontargeted}. FGSM and DeepFool are gradient-based attacks, that unsurprisingly need fewer predictions (queries). C\&W, a gradient-based attack as well, achieves the smallest perturbation with more queries. As decision-based attacks, qFool can get much smaller perturbation than Boundary attack. Besides, qFool has an obvious advantage in the number of iterations, that can bring significant speed-ups because of its high parallelization.

\begin{table*}[tp]
  \centering
  \fontsize{9}{10}\selectfont 
  \begin{threeparttable}
%   \vspace{-0.5}
  \caption{Comparison with different non-targeted attacks on the total number of backward gradient computation (\#gradients), the total number of forward predictions (\#predictions) and the total number of iterations (\#iterations). (FGSM$^{*}$ is a modified version of FGSM that tries to find the smallest perturbation in the direction of the sign of the gradient.)}
  \label{tab:comparison-nontargeted} 
    \begin{tabular}{ccccccc}
    \toprule
    &\multirow{2}{*}{\#gradients}&\multirow{2}{*}{\#predictions}&\multirow{2}{*}{\#iterations}&
    \multicolumn{3}{c}{MSE}\cr
    \cmidrule(lr){5-7} 
    &&&&VGG-19&ResNet-50&Inception-v3\cr
    \midrule
    FGSM$^{*}$&1&$\sim$69&$\sim$69&1.28e-6&6.37e-7&1.20e-5\cr
    DeepFool&$\sim$20&$\sim$20&$\sim$ 2&2.76e-7&1.71e-7&6.12e-7\cr
    C\&W&10000&10000&10000&2.22e-7&2.26e-7&3.25e-7\cr
    Boundary attack&0&$\sim$10000&$\sim$500&4.61e-4&1.15e-3&1.79e-3\cr
    \textbf{qFool-fullspace}&0&$\sim$ 10000&$\sim$3&1.16e-5&1.03e-4&2.48e-4\cr
    \textbf{qFool-subspace}&0&$\sim$ 10000&$\sim$3&6.15e-6&4.14e-5&1.80e-4\cr
    \bottomrule
    \end{tabular}
    \end{threeparttable}
    
\end{table*}

\begin{figure}
\centering
	\includegraphics[width=0.4\textwidth]{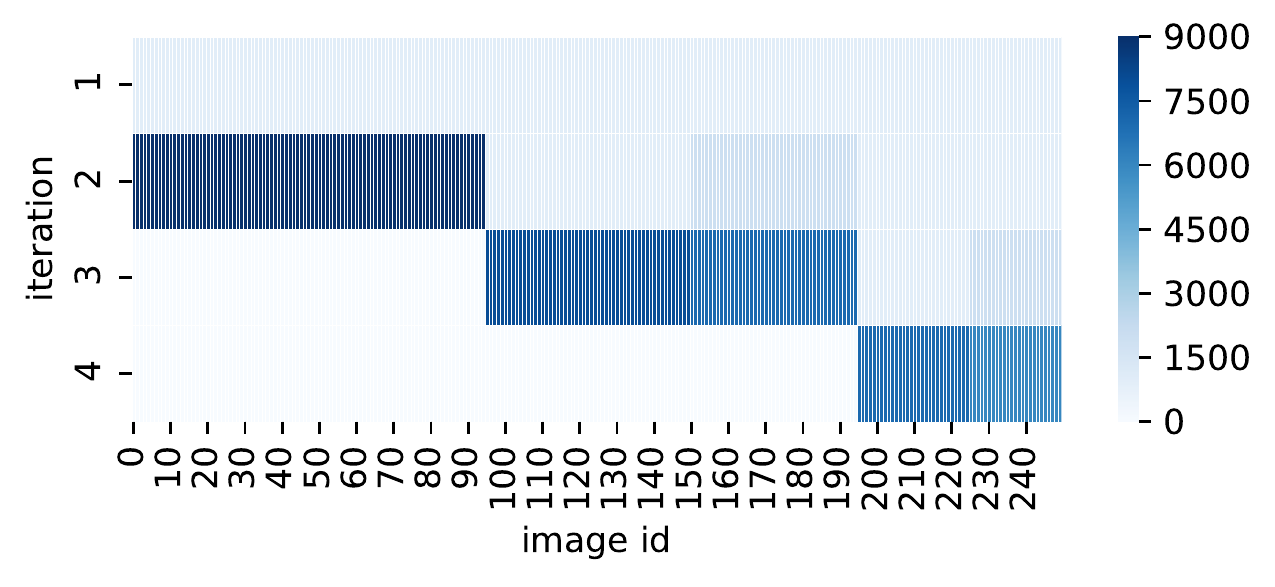}
	\vspace{-0.5em}
	\caption{Distribution of 10,000 queries in different iterations for 250 images. We have grouped the images according to the number of used iterations. More than 30\% images only need 2 iterations, and most queries appear in their second iteration. Other images need three or four iterations, and the last iteration similarly contain the most queries.} 
	\label{fig:iteration-query} 
\end{figure}

In addition, we observe in our experiments that, for 250 different images, the number of iterations is no more than 4, and the allocation of $n$ queries in consecutive iterations computed by our algorithm conforms to a similar rule: in the first few iterations, the numbers of queries are small, while the last iteration needs a much larger number of queries, shown in Fig. \ref{fig:iteration-query}. It indicates that the basic logic of qFool algorithm is reasonable and effective. Allocating some small numbers of queries to the first several iterations can make the starting point closer to the theoretical adversarial example, which leads to a more locally flat boundary at the starting point, 
and thus a more accurate approximation of adversarial perturbation direction.

\subsection{Targeted attacks}
\begin{figure*}[ht]
	\begin{minipage}[t]{0.32\textwidth}
		%\centering  
		\includegraphics[width=6cm]{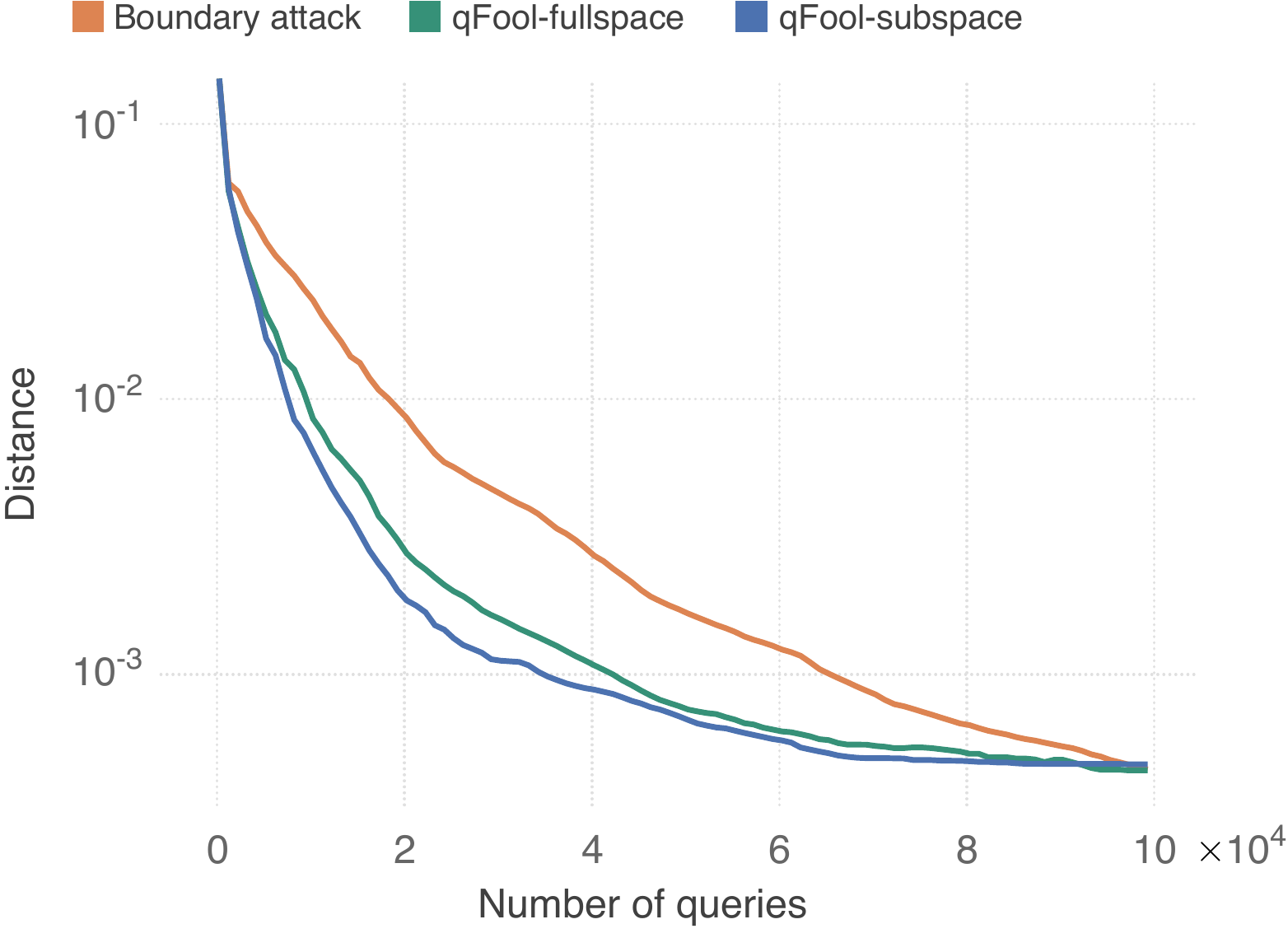}  
		\caption*{(a) VGG-19}
	\end{minipage} 
	\begin{minipage}[t]{0.32\textwidth}
		%\centering  
		\includegraphics[width=6cm]{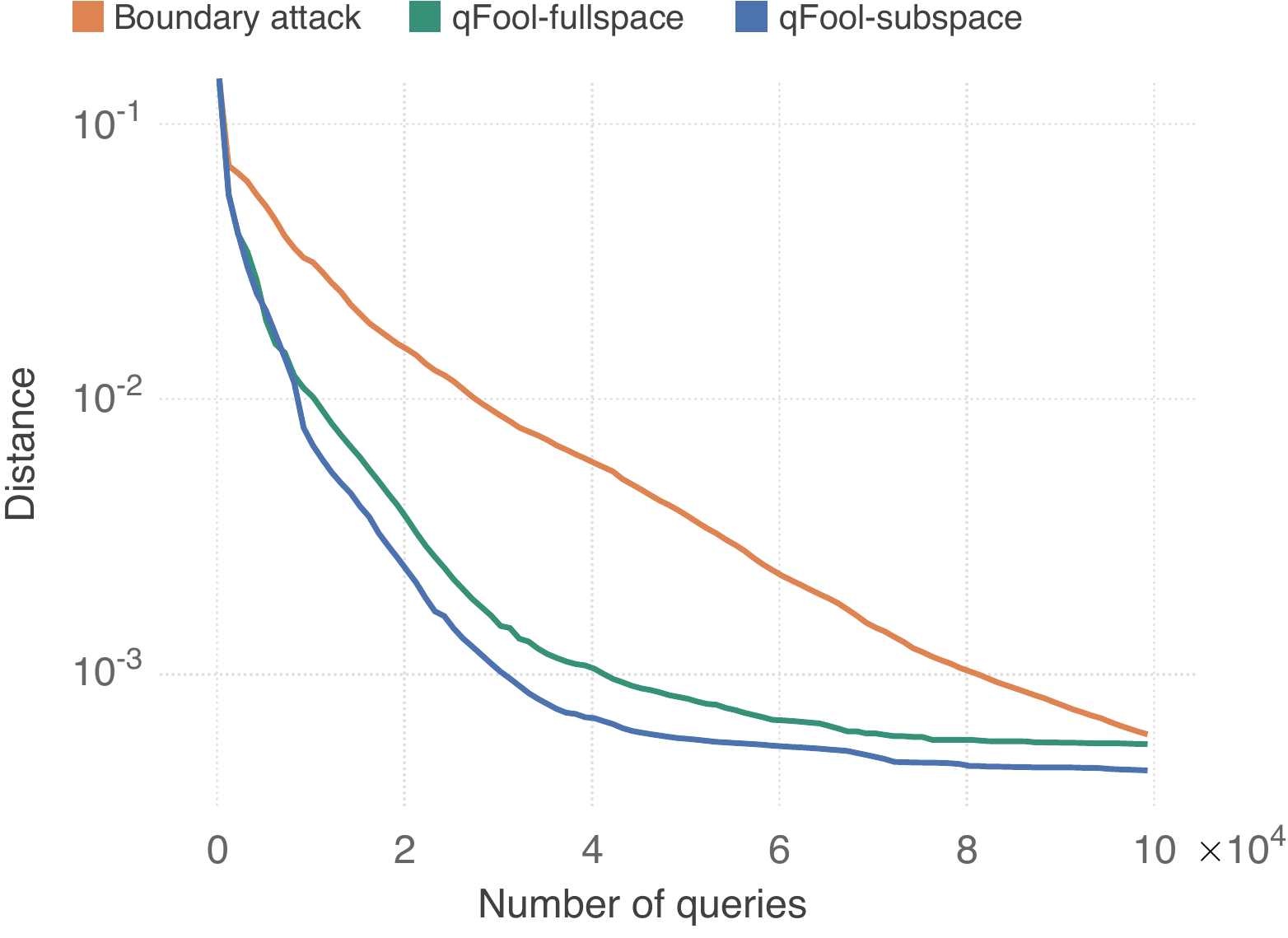}  
		\caption*{(b) ResNet-50}
	\end{minipage} 
	\begin{minipage}[t]{0.32\textwidth}
		%\centering  
		\includegraphics[width=6cm]{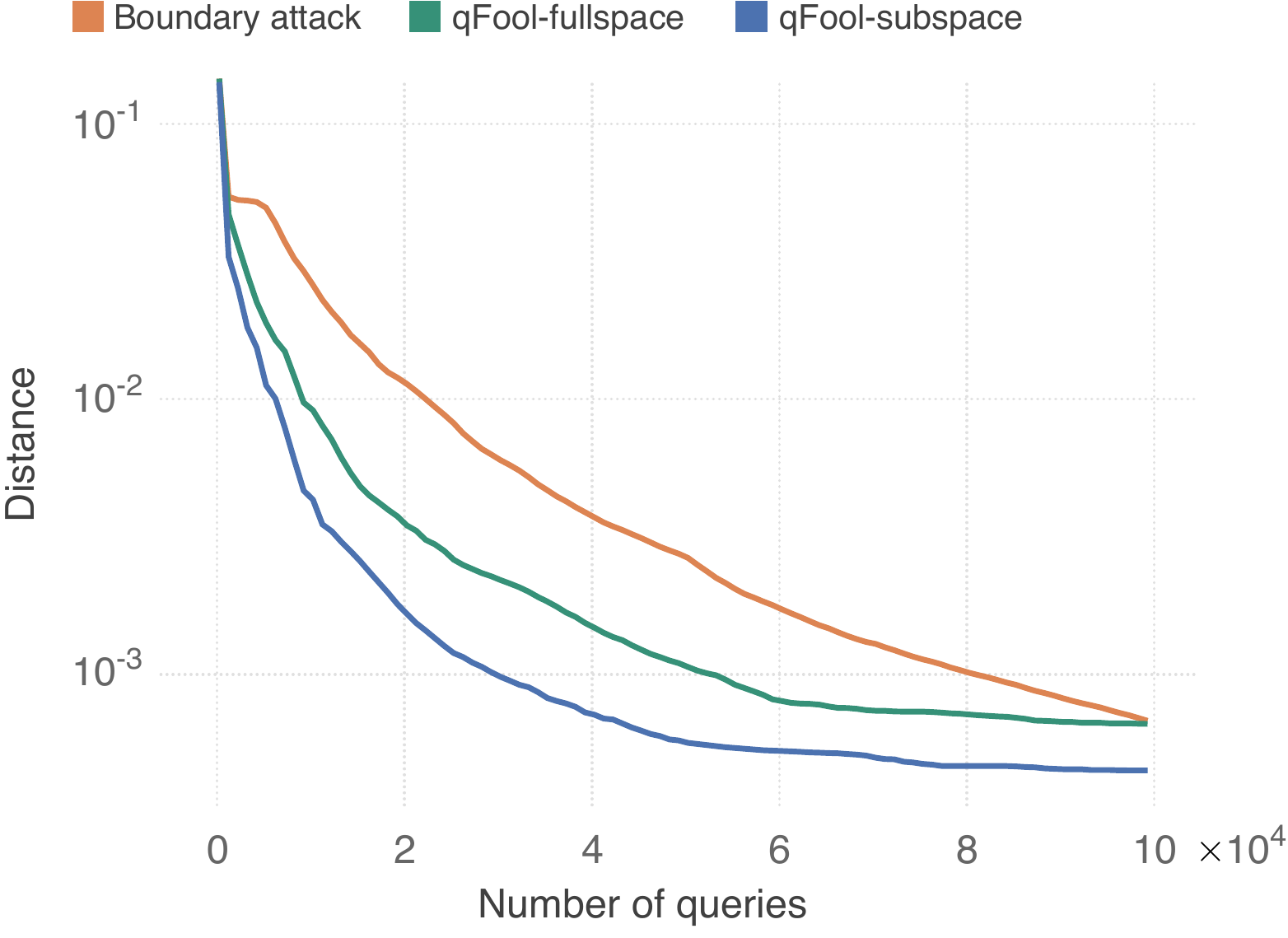}  
		\caption*{(c) Inception-v3}
	\end{minipage} 
	\vspace{-0.5em}
    \caption{Distance between adversarial example and original image versus numbers of queries in targeted qFool and Boundary attack \cite{brendel2017decision} in three different network models. }
    \label{fig:targeted}
\end{figure*}

\begin{figure*}[ht]
	\centering
	\begin{minipage}[t]{0.15\textwidth}
		\centering  
		\includegraphics[width=1.8cm]{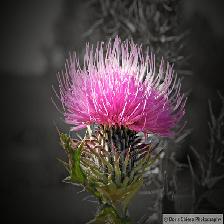}  
		\caption*{(a)}
	\end{minipage} 
	\begin{minipage}[t]{0.15\textwidth}
		%\centering  
		\includegraphics[width=1.8cm]{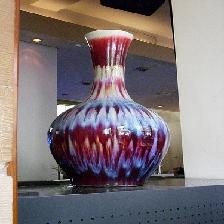}  
		\caption*{(b)}
	\end{minipage} 
	\begin{minipage}[t]{0.15\textwidth}
		%\centering  
		\includegraphics[width=1.8cm]{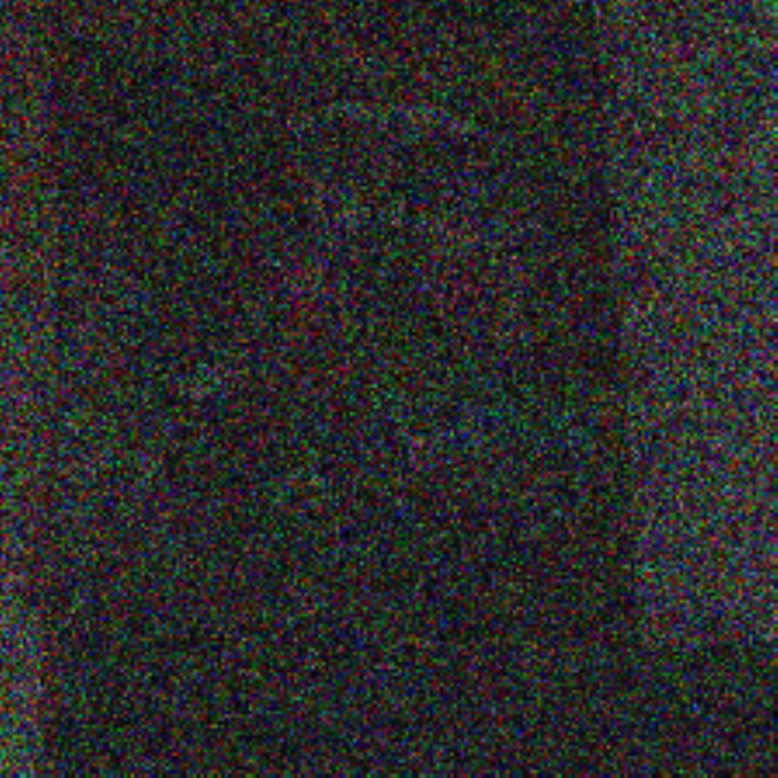}  
		\caption*{(c)}
	\end{minipage} 
	\begin{minipage}[t]{0.15\textwidth}
	%\centering  
	\includegraphics[width=1.8cm]{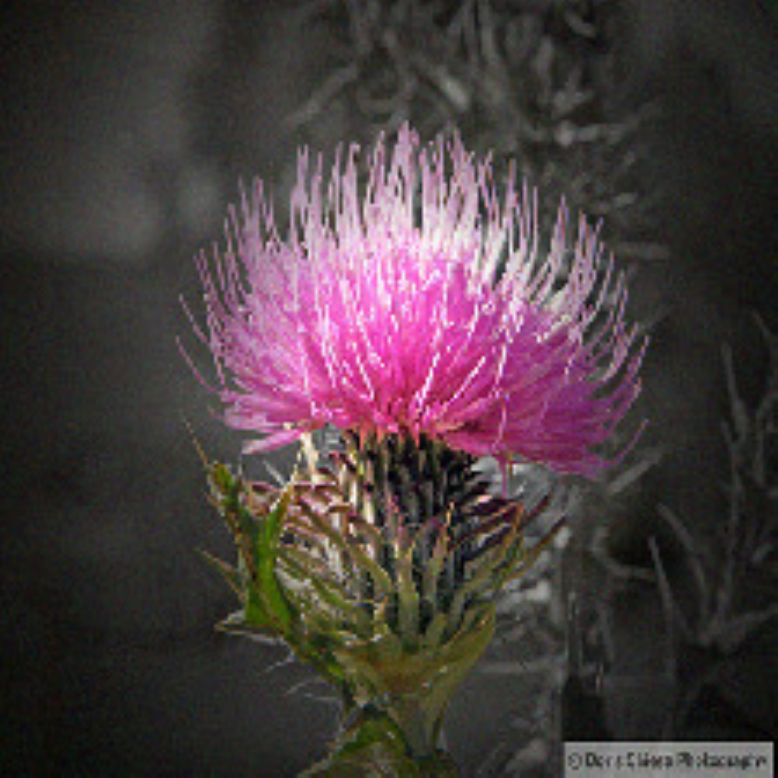}  
	\caption*{(d)}
\end{minipage} 
	\begin{minipage}[t]{0.15\textwidth}
	%\centering  
	\includegraphics[width=1.8cm]{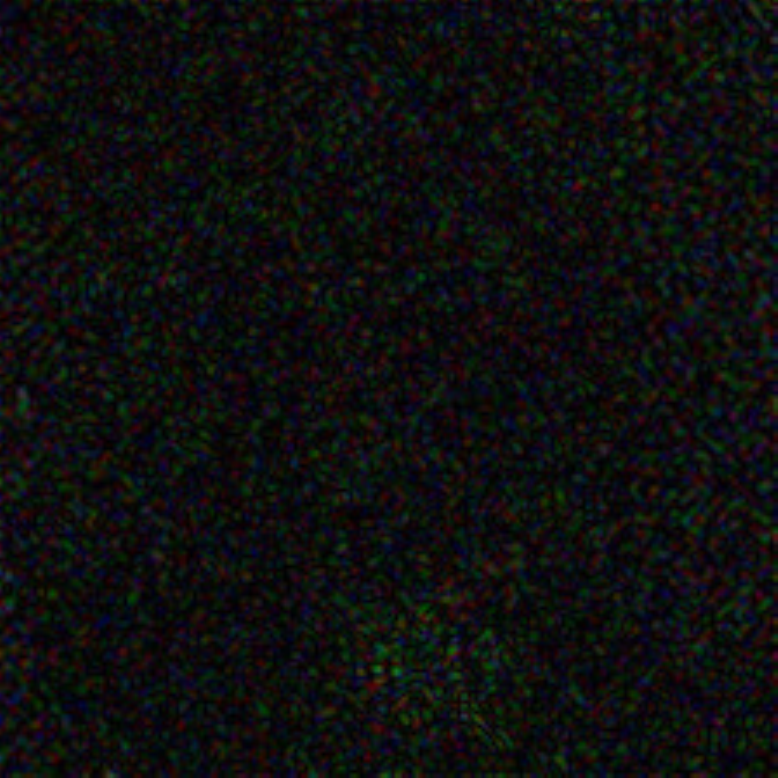}  
	\caption*{(e)}
\end{minipage} 
	\begin{minipage}[t]{0.15\textwidth}
	%\centering  
	\includegraphics[width=1.8cm]{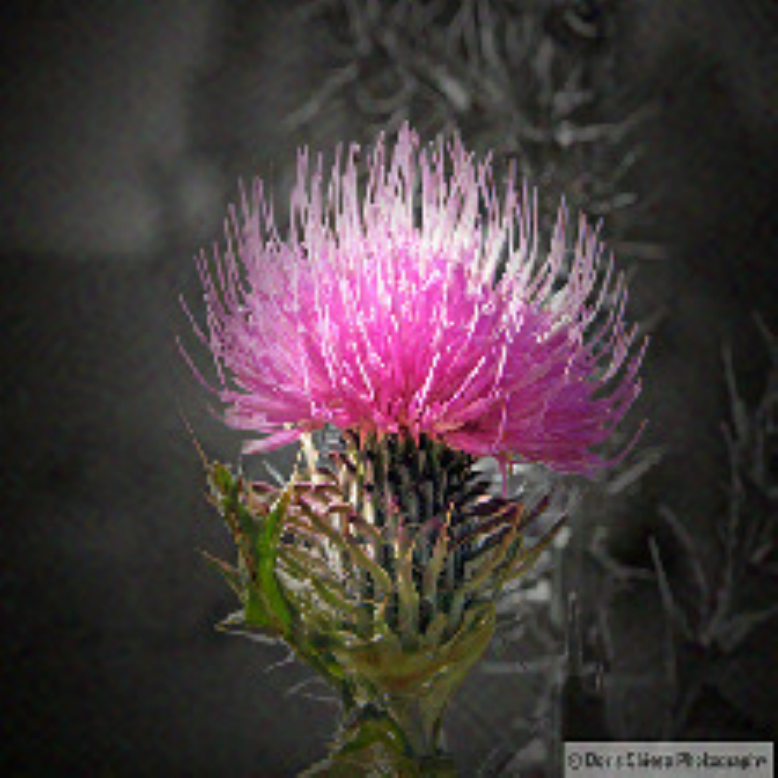}  
	\caption*{(f)}
\end{minipage} 
\vspace{-0.5em}
\caption{qFool in targeted attacks on VGG-19. (a) original:`cardoon' (b) target:`vase' (c) adversarial perturbation in full space (d) adversarial example in full space (e) adversarial perturbation in subspace (f) adversarial example in subspace.
	We use 50,000 queries in this experiment. The MSE is 4.40e-4 in full space and 1.12e-4 in subspace.}
	\label{fig: targeted-show}
\end{figure*}

In targeted attacks, we use 250 samples in ImageNet. For each of them, we draw a target label randomly and pick one target image belonging to the target label randomly. We keep this original-target image pair consistent in all experiments of targeted attacks. The results are shown in Fig.~\ref{fig:targeted}. We can see that, when the number of queries is not so large, our algorithm is more efficient than the Boundary attack method \cite{brendel2017decision}. Although the two curves cross at $100,000$ queries, qFool converges much faster. In the first several thousand queries, qFool can find a much smaller adversarial perturbation than Boundary attack. Therefore, in settings where the number of queries is limited, our algorithm is more efficient. Some samples are shown in Fig. \ref{fig: targeted-show}.

\begin{figure}[ht]
	\centering
		\begin{minipage}[t]{0.1\textwidth}
			%\centering  
			\includegraphics[width=1.5cm]{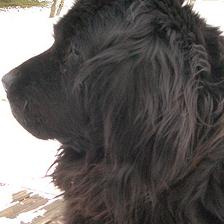}  
			\caption*{(a)}
		\end{minipage} 
		\begin{minipage}[t]{0.1\textwidth}
		\centering  
		\includegraphics[width=1.5cm]{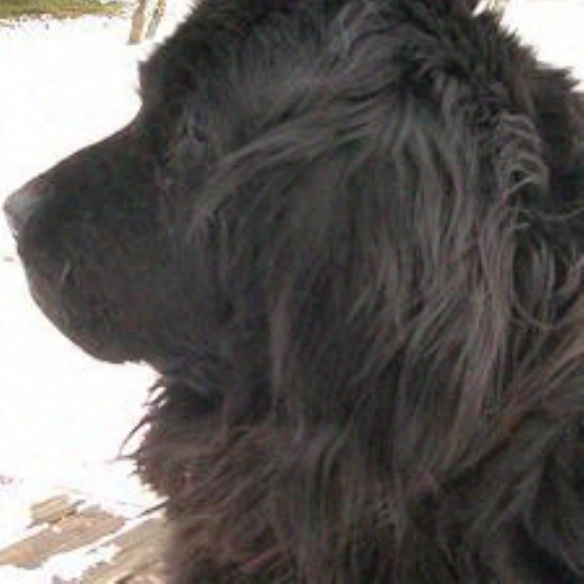}  
		\caption*{(b)}
	\end{minipage} 
	\begin{minipage}[t]{0.1\textwidth}
	%\centering  
	\includegraphics[width=1.5cm]{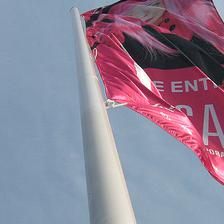}  
	\caption*{(c)}
\end{minipage} 
	\begin{minipage}[t]{0.1\textwidth}
	%\centering  
	\includegraphics[width=1.5cm]{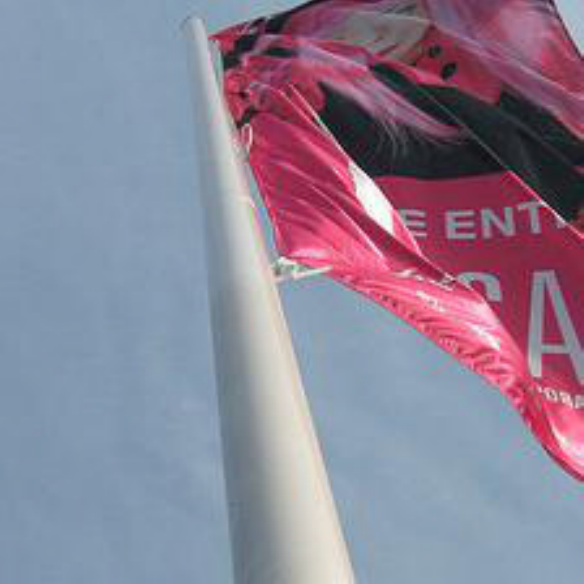}  
	\caption*{(d)}
\end{minipage} 
\vspace{-0.5em}
\caption{Attacking Google Cloud Vision. (a) original: `dog like mammal' (b) adversarial: `hair' (\textbf{1500} queries) (c) original: `flag' (d) adversarial: `sky' (\textbf{500} queries)}
\vspace{-1.5em}
\label{fig: attack-google}
\end{figure}

Different target labels or target images obviously result in different numbers of queries, as shown in Table 2. We list results at around 100,000 queries. It shows that MSE of adversarial examples is not positively correlated with the distance between the original and target image. Two images that are close in the spatial domain may not be similar in the feature domain.

\begin{figure*}[!h]
    \centering
	\includegraphics[width=16cm]{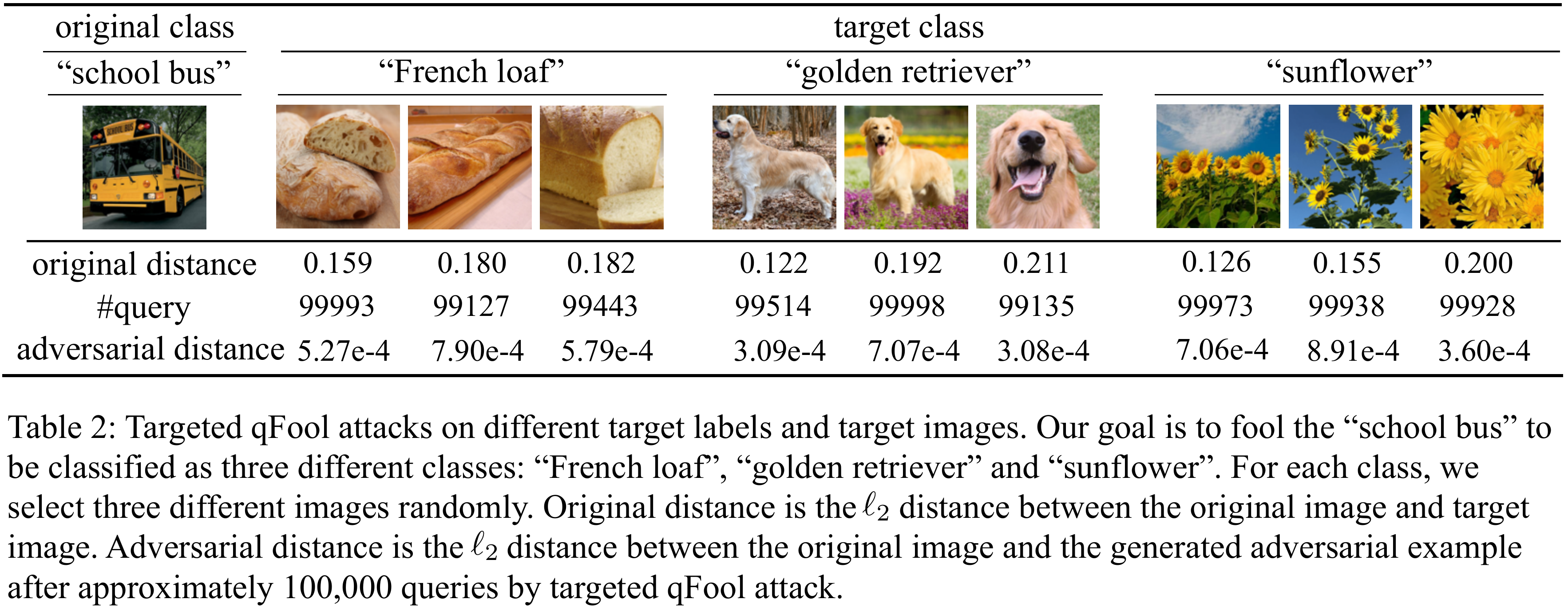}
	\vspace{-1em}
\end{figure*}

\subsection{Attacks on commercial classifiers}
 In order to demonstrate the effectiveness of qFool, we apply our non-targeted attack algorithm to one sample commercial system, namely Google Cloud Vision Image Recognition service, where users can get some tags associated with the input image and the corresponding confidence scores after querying the model. But we only use the final decision (top-1 tag) in order to attack this classifier. 
	
Adversaries would ideally want to use the smallest number of queries to get an adversarial example with a relatively good visual quality. Since qFool performs particularly well when the number of queries is not very large, it is suitable for attacking commercial classifiers. Fig.~\ref{fig: attack-google} demonstrates some adversarial examples generated by only $500\sim1500$ queries.  In the first `dog' image, the MSE is 7.53e-5, which is totally imperceptible by human eyes. We only use 1500 queries to the model. Its new top-1 label is `hair', and it indicates that the generated perturbation might mislead the model to ignore the whole content of this image yet focus on some local information. In the second `flag' image, we use only 500 queries and the MSE achieves 2.66e-6. The perturbation makes the model ignore the existence of the object flag, and classify the image as `sky'.

\section{Conclusion}
In this work, we propose a novel query-efficient decision-based attack algorithm, namely qFool, for settings where the attacker only has access to the final decision (top-1 label) of a model. Our algorithm is based on the observation that the curvature of the boundary is small around adversarial examples. We use a simple way to estimate the gradient direction of the boundary, and construct imperceptible adversarial examples using the estimated direction. With this method, the required number of queries in both our non-targeted and targeted attacks can be reduced drastically compared to previous decision-based attacks. We also enhance qFool by constraining the gradient estimation to a pre-defined subspace, which further reduces the number of queries. We finally apply qFool on Google Cloud Vision, demonstrating the effectiveness of our attack on a sample real-world black-box classifier.

\section*{Acknowledgements}
This work has been supported by a Google Faculty Research Award, and the Hasler Foundation, Switzerland, in the framework of the ROBERT project. We also gratefully acknowledge the support of NVIDIA Corporation with the donation of the GTX Titan X GPU used for this research.

{\small
\bibliographystyle{ieee}
\bibliography{egbib}
}

\end{document}